\documentclass[aps,prr,reprint,twocolumn,superscriptaddress,amsmath,amsfonts,amssymb]{revtex4-2}

\usepackage{bm}
\usepackage{graphicx}
\usepackage{subfigure}
\usepackage{array}
\usepackage{amssymb}
\usepackage{amsfonts}
\usepackage{amsmath}
\usepackage{mathrsfs}
\usepackage{color}
\usepackage{booktabs}
\usepackage{threeparttable}
\usepackage{multirow}
\usepackage{times}
\usepackage{epsfig}
\usepackage{threeparttable}
\usepackage{chngpage}
\usepackage{latexsym}
\usepackage[citecolor=blue,linkcolor=blue,colorlinks=true]{hyperref}%
\usepackage{changes}
\usepackage{ulem}

\newtheorem{theorem}{Theorem}

\begin{document}
 \title{ Embedding Theory of Reservoir Computing and Reducing Reservoir Network Using Time Delays}

 \author{Xing-Yue Duan}
\affiliation{School of Mathematical Sciences, Soochow University, Suzhou 215006, China}

\author{Xiong Ying}
\affiliation{School of Mathematical Sciences, SCMS, and SCAM, Fudan University, Shanghai 200433, China}

\author{Si-Yang Leng}
\affiliation{Research Institute of Intelligent Complex Systems and Centre for Computational Systems Biology, Fudan University, Shanghai 200433, China}
\affiliation{Academy for Engineering and Technology, Fudan University, Shanghai 200433, China}
\affiliation{State Key Laboratory of Medical Neurobiology, LCNBI, and MOE Frontiers Center for Brain Science, Institutes of Brain Science, Fudan University, Shanghai 200032, China}

\author{J\"{u}rgen Kurths}
\affiliation{Research Institute of Intelligent Complex Systems and Centre for Computational Systems Biology, Fudan University, Shanghai 200433, China}
\affiliation{Potsdam Institute for Climate Impact Research (PIK), Potsdam 14473, Germany}

\author{Wei Lin}\email{wlin@fudan.edu.cn}
\affiliation{School of Mathematical Sciences, SCMS, and SCAM, Fudan University, Shanghai 200433, China}
\affiliation{Research Institute of Intelligent Complex Systems and Centre for Computational Systems Biology, Fudan University, Shanghai 200433, China}
\affiliation{State Key Laboratory of Medical Neurobiology, LCNBI, and MOE Frontiers Center for Brain Science, Institutes of Brain Science, Fudan University, Shanghai 200032, China}
\affiliation{Shanghai Artificial Intelligence Laboratory, Shanghai 200232, China}

\author{Huan-Fei Ma}\email{hfma@suda.edu.cn}
\affiliation{School of Mathematical Sciences, Soochow University, Suzhou 215006, China}


\date{\today}

\begin{abstract}
Reservoir computing (RC), a particular form of recurrent neural network, is under explosive development due to its exceptional efficacy and high performance in reconstruction or/and prediction of complex physical systems.  However, the mechanism triggering such effective applications of RC is still unclear, awaiting deep and systematic exploration.  {Here,
combining the delayed embedding theory with the generalized embedding theory, we rigorously prove that RC is essentially a high dimensional embedding of the original input nonlinear dynamical system.   Thus, using this embedding property, we unify into a universal framework the standard RC and the time-delayed RC where we novelly introduce time delays only into the network's \textit{output layer}, and we further find a trade-off relation between the time delays and the number of neurons in RC.  Based on these findings, we significantly reduce the RC's network size and promote its memory capacity in completing systems reconstruction and prediction.  More surprisingly, only using a \textit{single-neuron} reservoir with time delays is sometimes sufficient for achieving reconstruction and prediction tasks, while the standard RC of any large size but without time delay cannot complete them yet.}
\end{abstract}


\maketitle

The last decades have witnessed the extensive application and development of machine learning technology in data-driven research and in high-technology-oriented industry as well.  As a representative leader among many machine learning techniques, the artificial neural network (ANN) has emerged as a powerful approach that is well suited for coping with the supervised learning problems.  Among various architectures of ANN,  Reservoir Computing (RC), which is a recently developed framework \cite{jaeger2004}, a special variant of a recurrent neural network, and also known as a generalization of echo-state network (ESN) \cite{jaeger2001} or liquid state machine (LSM) \cite{maass2002}, has been reported to
have great efficacy in reconstruction or/and prediction of many complex physical systems only based on the observational data of time series \cite{PRL2018,chaos2019upo,tanaka2019,chaos2020review}.
The architecture of RC is quite contracted.  As shown in Fig.~\ref{fig1a}, only three weight matrices are involved: the input matrix and the reservoir recurrent matrix are randomly generated but fixed, while the output matrix is determined via training.   As such,  efficient least squares optimization methods rather than the resource-consuming back propagation algorithm are adopted in the training process \cite{jaeger2002tutorial}. {Behind such a contracted architecture, two questions arise naturally: ``What is the fundamental mechanism resulting the efficacy of RC?'' and ``How to improve the structure using the uncovered mechanism?''  These questions have attracted great attention and motivated abundant discussions, }
including those from the topology and the complexity of random connections \cite{topology2019,connectivity2019} {to the spectral radius of random networks and the edge of chaos \cite{jaeger2009survey,Lai2019,boedecker2012information}}, from the fading memory property \cite{fadmemory2012} to the echo state property \cite{JaegerESP2012,jaegerESP2013}, from the choice of activation functions \cite{sigmoidshape2019} to the training algorithm of the output layer \cite{sigmoidlearning2007}.  {Yet}, recent understanding of RC is often via heuristic interpretation and it is widely believed that a successful RC should possess high dimensionality, nonlinearity, fading memory, and separation property \cite{tanaka2019}, but barely with rigorous and mathematical demonstrations.

In order to decipher the RC's capacity of reconstructing and forecasting nonlinear dynamics, several efforts from a viewpoint of dynamical systems have been recently made.  For example, {the regression model and the dynamical model decomposition method were used to illustrate the usefulness of RC to forecasting chaotic dynamics \cite{chaos2021,NC2021}, and, to demonstrate the approximation capability of RC, an embedding conjecture was studied and could be partially validated for a specific form of RC under right technical conditions   \cite{NN2020,hart2021echo}.
In the area of photonic neural network, an architecture of photonic reservoir computing has been developed through using a spatiotemporal analogy to translate a delayed differential equation (DDE) into a virtual single-neuron reservoir network \cite{nc2011,brunner2018tutorial,PRX2017}.
Still, despite these significant efforts and achievements, some key questions remain unsolved: ``How to understand the network dimension of general RC using the theories of nonlinear dynamics and functional analytics?'' and ``How to design a small size network in RC for sustaining its efficacy?''

\begin{figure}
    \centering
    \subfigure{\label{fig1a}}\subfigure{\label{fig1b}}\subfigure{\label{fig1c}}
    \includegraphics[width=0.47\textwidth]{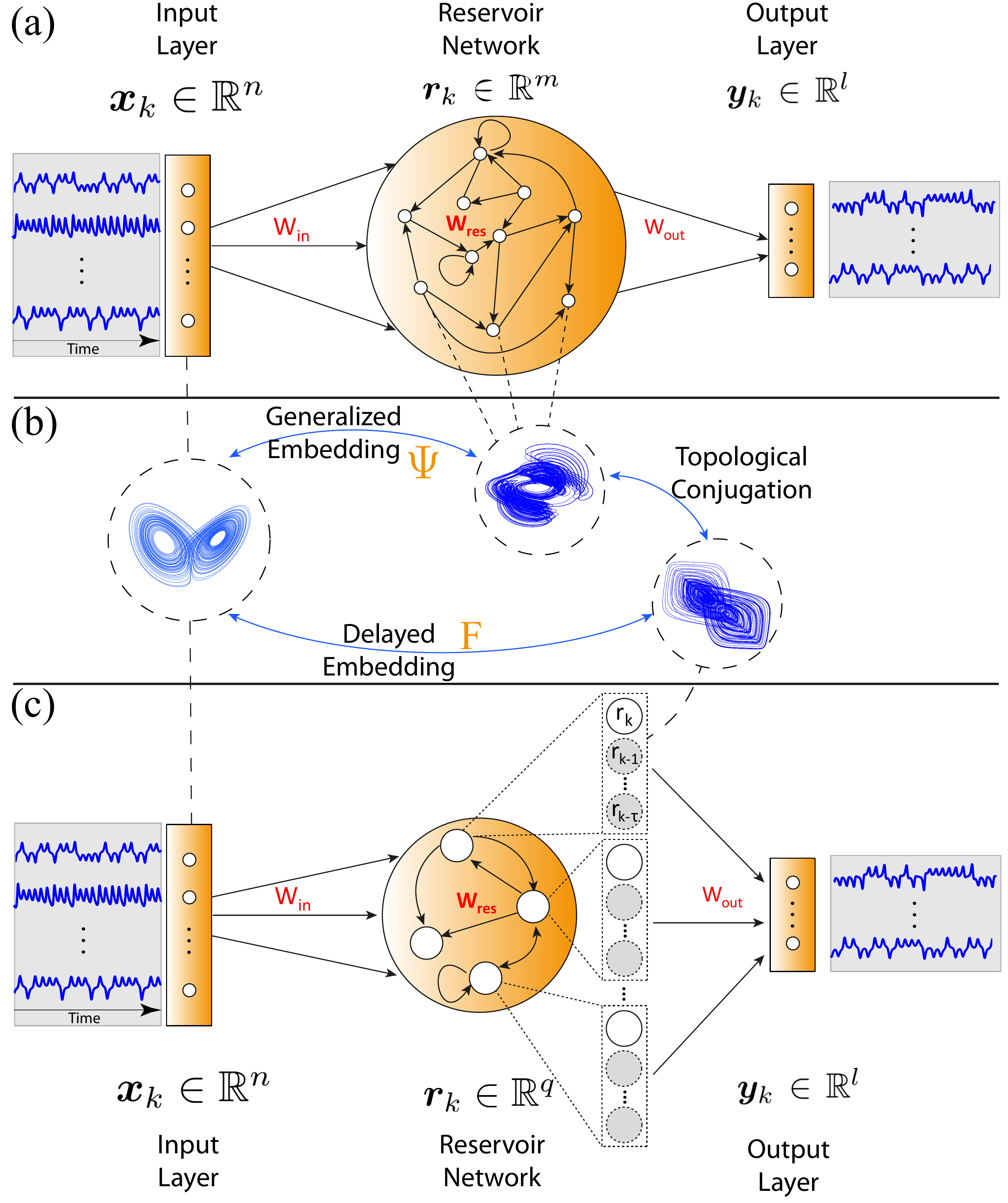}
    \caption{\label{fig1} RCs as different nonlinear dynamical systems and embeddings. (a) A standard RC without time-delay. (b) {The generalized embedding $\Psi$ from the input dynamics to the standard non-delayed reservoir network and the delayed embedding $F$ from the input dynamics to the delayed reservoir network, which constitute a topological conjugation between the dynamics of the non-delayed reservoir network and the delayed reservoir network.} (c) A time-delayed RC with a smaller network size in the reservoir layer.}
\end{figure}

In this Letter, we rigorously study the mechanism of RC from a viewpoint of nonlinear dynamical systems and novelly propose a framework of time-delayed RC.  Particularly combining the delayed embedding theory with the generalized embedding theory, we first prove a general reservoir network rigorously as a high dimensional embedding of the original input nonlinear dynamical system.  Then, we further reveal a trade-off relation between the time delays and the number of neurons by unifying into a universal framework the standard RC without delays and the time-delayed RC where the time delays are introduced into the network's output layer.  It therefore allows us to construct a random reservoir network with a significantly-reduced physical dimension to achieve the efficacy that the original larger-size RC owns.
Surprisingly, we show that a standard reservoir of single-neuron, without introducing any DDE or time-division multiplexing technique, can sometimes work well for reconstructing and forecasting some representative physical systems.  Moreover, we find flexible memory capacity in the time-delayed RC, which makes it possible to accomplish more challenging tasks of dynamics reconstruction that cannot be easily achieved using a standard RC of the same scale.

We start with a standard RC as sketched in Fig.~\ref{fig1a}. Here, the input data $\bm{x}_k\in\mathbb{R}^n$ represents the state vector of a dynamical system that is evolving on a compact manifold $\mathcal{M}$ with the evolution operator $\varphi\in {\rm Diff}^2(\mathcal{M}): \bm{x}_{k+1}=\varphi(\bm{x}_k)$.  The vector $\bm{r}_k\in\mathbb{R}^m$ represents the state of $m$ reservoir neurons at time step $k$, the input layer weight matrix $W_{{\rm in}}$ and the reservoir network matrix $W_{{\rm res}}$ are, respectively, $m\times n$ and $m\times m$ random matrices generated according to certain distribution laws. The dynamical evolution of the reservoir neurons is governed by (RN):
$
  \bm{r}_k=(1-\alpha)\bm{r}_{k-1}+\alpha \phi(W_{{\rm res}}\bm{r}_{k-1}+W_{{\rm in}}\bm{x}_k),
$
where $\alpha$ is the leakage factor, and $\phi\in C^2(\mathbb{R},(-1,1))$ is set a sigmoid function (e.g., $\tanh$) in this Letter. The output vector $\bm{y}_k\in\mathbb{R}^l$ is determined by the output weight matrix $W_{\rm out}\in\mathbb{R}^{l\times m}$ such that $\bm{y}_k=W_{\rm out}\bm{r}_k$. In the task of nonlinear system reconstruction, given the time series, denoted by $\bm{x}_k,k=1,\cdots,N+1$, as training data, the target is to train the output weight matrix $W_{\rm out}$ so as to approximate the one-step dynamics prediction, i.e., $\bm{y}_k\approx\bm{x}_{k+1}$. To achieve this, the output weight matrix $W_{\rm out}$ is generally calculated by minimizing the loss function
$
  \mathcal{L}=\sum_{k=1}^N\|\bm{x}_{k+1}-W_{\rm out}\bm{r}_k\|^2+\beta\|W_{\rm out}\|^2
$
over the training data set, where $\beta>0$, the $L_2$-regularization coefficient, is introduced to make optimization robust.
  After training, one can fix the output weight matrix $W_{\rm out}$ and redirect the output $\bm{y}_{k}=W_{\rm out}\bm{r}_k$ as an approximation of $\bm{x}_{k+1}$ into the input layer of the network and thus generate the autonomous dynamics for $\bm{x}_k$ with $k>N$.

To rigorously establish an embedding theory for RC, we consider directly the evolution (RN) of the reservoir neurons with the leakage factor $\alpha=1$ as:
$$
\begin{array}{l}
\bm{r}_{k+1,\bm{x}_0}^{\bm{b}_0}=\phi(W_\textrm{res}\bm{r}_{k,\bm{x}_0}^{\bm{b}_0}+W_\textrm{in}\varphi^{k+1}\bm{x}_0), ~~ k=0,1,\cdots,
\end{array}
$$
 and define a map as $\mathfrak{G}^k[\bm{r}_{0},W_\textrm{res},W_\textrm{in}](\bm{x}_0)=\bm{r}_{k,\bm{x}_0}^{\bm{b}_0}$.  Here, $\bm{r}_{0,\bm{x}_0}^{\bm{b}_0}=\bm{b}_0$, $\bm{b}_0\in\mathbb{I}^m$,  and $\mathbb{I}=(-1,1)$.
 Thus, we rigorously have the following result.

\begin{theorem}\label{theom}
  Let $m\geqslant 2\textrm{\rm dim}(\mathcal{M})+1$ and $[\bm{r}_{0},W_\textrm{\rm res},W_\textrm{\rm in}]\in\mathbb{I}^m\times \mathbb{R}^{m\times m}\times\mathbb{R}^{m\times n}$ with
  $\mbox{\rm dim}(\mathcal{M})$ as the box-counting dimension of the manifold $\mathcal{M}$.   Then, there exists a number $k^*>0$, such that $\mathfrak{G}^k[\bm{r}_{0},W_\textrm{\rm res},W_\textrm{\rm in}]\in C^1(\mathcal{M},\mathbb{R}^{m})$ is generically an embedding for all $k>k^*$.
\end{theorem}

Here, the generic conclusion in Theorem~\ref{theom} means that, for all $[\bm{r}_{0},W_\textrm{res},W_\textrm{in}]\in\mathcal{S}$ where $\mathcal{S}\subset\mathbb{I}^m\times \mathbb{R}^{m\times m}\times\mathbb{R}^{m\times n}$ is an open and dense set,
$\mathfrak{G}^k[\bm{r}_{0},W_\textrm{res},W_\textrm{in}]$ is an embedding for any sufficiently large $k$. The detailed and rigorous proof with respect to the $C^{1}$-topology
is provided in Supplemental Information (SI) \cite{SM}.
Moreover, the echo state property, a necessary condition for constructing an RC,  requires that, with the general configuration $\{W_{\rm in},W_{\rm res},\phi\}$, the evolutions (RN) of the reservoir neurons, starting from any different initial values $\bm{r}^{(1)}_0$ and $\bm{r}^{(2)}_0$, converge to the same dynamics, i.e., $\lim_{k\rightarrow\infty}\|\bm{r}^{(1)}_k-\bm{r}^{(2)}_k\|=0$ \cite{JaegerESP2012}.
Hence, by virtue of Theorem~\ref{theom}, regardless of the choice of the initial value $\bm{r}_0$, the dynamics of reservoir neurons is determined by the input dynamics, i.e.,  {\it there exists a unique embedding $\Psi$ such that $\bm{r}_k=\Psi(\bm{x}_k)$ after a transient phase, while each component $r_{ik}=\Psi_i(\bm{x}_k)$ implies that the dynamics of each neuron is an observable of the original dynamics.}

\begin{figure}
    \centering\includegraphics[width=.4\textwidth]{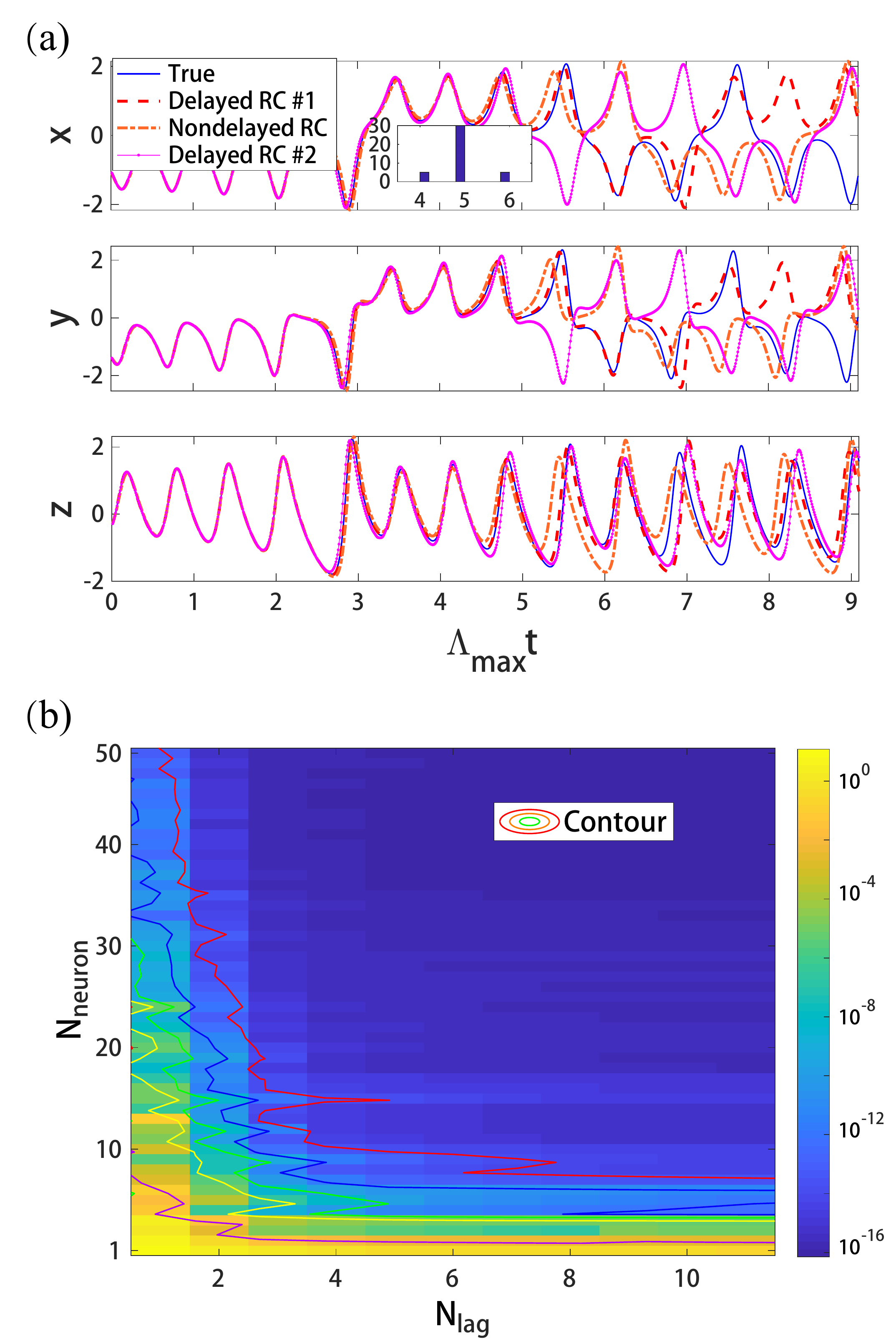}\subfigure{\label{fig2a}}\subfigure{\label{fig2b}}
    \caption{\label{fig2} (a) Reconstructed dynamics of the Lorenz system by a non-delayed RC including $200$ neurons, a delayed RC \#1 including $40$ neurons with uniformly $5$ lags for each neuron, and a delayed RC \#2 including $40$ neurons with random lags for each neuron. Here, the time unit is expressed in the Lyapunov time, and the random lags are generated by a distribution centered at $5$ as shown in the inset. (b) System reconstruction test for the Lorenz system with different combinations of $N_{\rm neuron}$ and $N_{\rm lag}$, where the training MSE in a log-scale and the contour curves are, respectively, highlighted. {Here, $\tau=5$ and the sampling stepsize is $\Delta t=0.01$. All the other parameter settings are introduced in \cite{SM}.}   }
\end{figure}

In the standard RC investigated-above, $m$, the number of reservoir neurons and also
known as the reservoir dimension, is often required to be huge \cite{jaeger2002tutorial,tanaka2019}.
To design a different RC framework, significantly reducing $m$, we introduce time delays into the output layer, as sketched in Fig.~\ref{fig1c}.  While all the configuration $\{W_{\rm in},W_{\rm res},\phi\}$ and the input data $\bm{x}$ are set in the same manners, the reservoir network is assumed to include $q~(<m)$ neurons only.  Thus, a new reservoir vector before the output layer is designated as
$
\tilde{\bm{r}}_k=[r_{1,k},r_{1,k-\tau},\cdots,r_{1,k-d_1\tau+\tau},\cdots,r_{q,k}, \cdots,r_{q,k-d_q\tau+\tau}]^\top,
$
and, correspondingly, the output matrix $W_{\rm out}$  is calculated by minimizing the $L_2$ loss function
$$
  \tilde{\mathcal{L}}=\sum_{k=1}^{N}\|\bm{x}_{k+1}-W_{\rm out}\tilde{\bm{r}}_k\|^2+\beta\|W_{\rm out}\|^2
$$
with $W_{\rm out}\in\mathbb{R}^{l\times d}$ and $d=\sum_{i=1}^{q}d_i$.
Here, the new reservoir vector $\tilde{\bm{r}}_k$ is formed by the lagged dynamics of each neuron, i.e., $q$ neurons with each neuron contributing $d_i$ lagged dynamics $[r_{i,k},r_{i,k-\tau},\dots,r_{i,k-d_i\tau+\tau}]$ where $\tau$ is a time delay.  Assigning $d$ as the output dimension of this delayed RC.

Now, we are in a position to demonstrate that the time-delayed RC with the above-assigned $d$ has the same representation and computation ability as the standard RC} involving $m$ neurons without time delay under the same parameter settings, as long as $d\thickapprox m$.   Actually, based on the delayed embedding theory and its applications \cite{takens,embedology,PNAS2018,NSR2022}, an approximate combination of the lagged observable can also generically form an embedding, i.e., for smooth observational functions $\Psi_1,\cdots,\Psi_q$,
$
F(\bm{x})=\big[\Psi_1(\bm{x}),\Psi_1(\varphi^{-1}(\bm{x})),\cdots,\Psi_1(\varphi^{-(d_1-1)}(\bm{x})),\cdots,
\Psi_q(\bm{x}),  \\
 \Psi_q(\varphi^{-1}(\bm{x})),\cdots,\Psi_q(\varphi^{d_q-1}(\bm{x}))\big]
$
is generically an embedding as long as $\sum_{i=1}^{q}d_i>2\textrm{dim}(\mathcal{M})$. Using the above-obtained conclusion that each neuron is generally an observable, we further conclude that the proposed new reservoir vector $\tilde{\bm{r}}_{k}$ is also an embedding.
Thus, the dynamics of the state vector ${\bm{r}}_{k}$ in the  $m$-neuron reservoir network without time delay is topologically conjugated with the dynamics of the reservoir vector $\tilde{\bm{r}}_{k}$ of a $q$-neuron reservoir network in the sense of embedding as long as $m=d$ with $d=\sum_{i=1}^{q}d_i$, as sketched in Fig.~\ref{fig1b}.  Consequently, we come to a conclusion that {\textit{the delayed observables of the RC state, seen as additional nonlinear observables, have the same computational power in the system reconstruction. }

To demonstrate the capability of our time-delayed RC, we first consider the benchmark Lorenz system.  After a training phase including $N=6,000$ samples, the autonomously-generated dynamics by the RC are shown in Fig.~\ref{fig2a}. Particularly, used are a standard RC,
a time-delayed RC containing fewer neurons with uniformly lagged dynamics for each neuron, and a time-delayed RC containing the same number of neurons but with random lags for each neuron.
Clearly, the time-delayed RC has almost the same performance of system reconstruction as the non-delayed one, no matter the lags are uniformly or randomly generated.
Actually, this coincides with the above-performed arguments from a viewpoint of embedding that the dynamics of this non-delayed RC  is a generalized embedding to the input dynamics with generically $200$ observables, while the dynamics of the time-delayed RC  forms an embedding of dimension $200$ when the sum of lags equals $200$ for either uniform or random lags. Such a trade-off relation is further clearly illustrated in Fig.~\ref{fig2b} where different neuron number with different lag number for each neuron is combined, and, for each combination, a training error is calculated as the mean squared error (MSE) on the training data set based over $20$ independent runs.
As depicted in Fig.~\ref{fig2b},  for a fixed moderate number of neurons, the training error decreases monotonically with the lag number for each neuron, and, for a fixed moderate lag number, the training error also decreases monotonically with the neuron number. Analogous results are also obtained for the other benchmark systems, as presented in \cite{SM} (see Fig.~S5). All these further reinforce the above conclusion that, whenever $d\thickapprox m$ with moderate $N_{\rm lag}$ and $N_{\rm neuron}$, the time-delayed and the non-delayed RCs generally share the same ability in system representation, and the numbers of neurons and of time lags can be traded off mutually in these frameworks.

Such a trade-off relationship further puts the non-delayed and time-delayed RCs into a unified framework where the output dimension $d$ becomes the effective reservoir dimension that finally decides the ability of the system representation. The standard non-delayed RC is actually a degenerated form in this unified framework where all the neurons have zero lag.  More surprisingly, we find that it is even possible to reduce the number of neurons into one and realize a single-neuron reservoir in the proposed framework. To see this, we consider a gene regulation model with multiple delays: $\dot{x}(t)=-kx(t)+gf_1(x(t-\tau_1))f_2(x(t-\tau_2))$ which describes self-inhibitory and self-activation with distinct delays $\tau_1$ and $\tau_2$, and with specific parameters the one-dimensional model has chaotic dynamics \cite{suzuki2016periodic,wernecke2019chaos,SM}.  {Specifically, a time-delayed RC including only one neuron with $600$ lags is used to reconstruct the dynamics, and the autonomously generated dynamics after training are shown in Fig.~\ref{fig3a}. The results confirm that the single-neuron, time-delayed RC performs well, achieving the same reconstruction ability of the time-delayed RC with multiple neurons.
Frankly, the single-neuron RC in this numerical illustration is only a special case that is not universally suitable for any system reconstruction.  Due to multi-scale property, the task of system reconstruction for multi-variable using one reservoir network usually requires more than one single-neuron.}   As for the task in Fig.~\ref{fig2a}, in order to get a successful prediction for the $3$ components of the Lorenz system, a single-neuron reservoir is not adequate even with multiple time delays.
\begin{figure}
    \centering
    \subfigure{\label{fig3a}}\subfigure{\label{fig3b}}
    \includegraphics[width=0.4\textwidth]{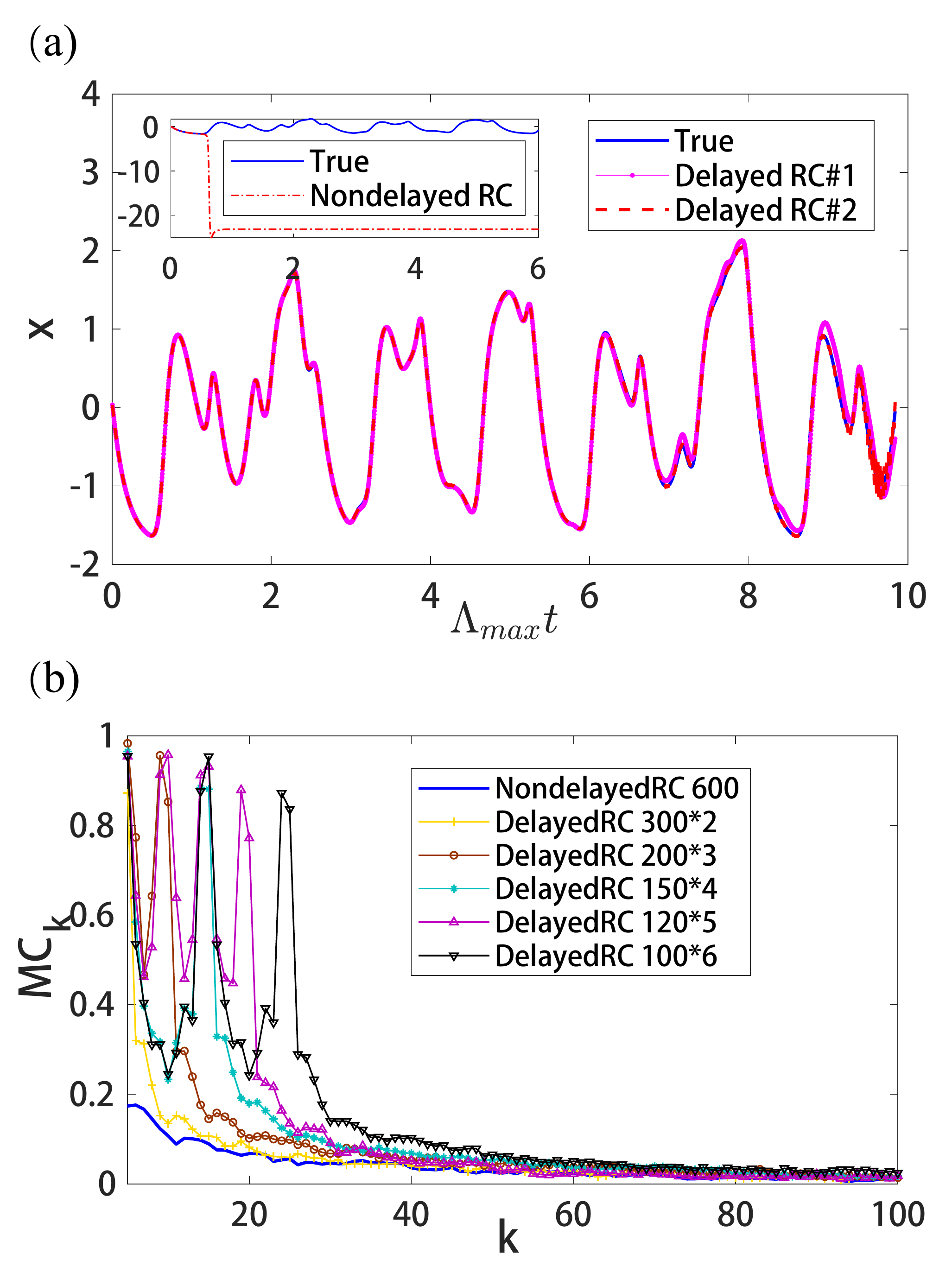}
    \caption{{ (a) Reconstructed dynamics of the chaotic gene regulation model by the standard RC including 600 neurons (see the inset), the single-neuron, time-delayed RC\#1 with 600 lags, and the time-delayed RC\#2 including 6 neurons and 100 lags for each neuron. (b) MC test with different combinations of $N_{\textrm{neuron}}\cdot N_{\textrm{lag}}$. Here, $\tau=5$ and the sampling stepsize is $\Delta t=0.1$.  All the other parameter settings are introduced in \cite{SM}. } }
    \label{fig3}
\end{figure}
 In addition to the equivalent representation ability in the sense of embedding, we further discover that the time-delayed RC has a more flexible memory capacity which is an essential measure for RC's reconstruction ability for delayed systems.
 In the dynamics reconstruction job for the above gene regulation model in Fig.~\ref{fig3a}, the chaotic dynamics cannot be reconstructed by a standard RC, no matter how large the reservoir is, according to the dimension test \cite{SM}.  However, with all the same reservoir environment, the time-delayed RC [both RC\#1 and RC\#2 in Fig.~\ref{fig3a}] can fulfill the job quite well. To understand this phenomenon, we calculate the memory capacity (MC) for different RC frameworks, using the definition in \cite{jaeger2002tutorial} and with different combinations of $N_{\textrm{neuron}}$ and $N_{\textrm{lag}}$ but satisfying the same output dimension, i.e., $N_{\textrm{neuron}}\cdot N_{\textrm{lag}}=600$.  Specifically, MC of a reservoir refers to its ability to retain information from previous time steps and it is defined in \cite{jaeger2002tutorial} as
  \begin{align*}
	\label{eq:MCk}
	{\rm MC}_k=\frac{\operatorname{cov}\left(x(t-k), \hat{y}_{k}(t)\right)^2}{\operatorname{var}(x(t)) \cdot \operatorname{var}\left(\hat{y}_{k}(t)\right)},
\end{align*}
where a random sequence of input values $x(t)$ is presented to the reservoir, and the reservoir output $\hat{y}_k(t)$ is trained to predict a previous input value $x(t-k)$, and here $\text{cov}(\cdot)$ and $\text{var}(\cdot)$, respectively, represent covariance and variance.

 Figure~\ref{fig3b} clearly shows that, as $N_{\textrm{lag}}$ increases, the reservoir computer with different delay settings has stronger memory capacity though still keeping a fading memory fashion.  This is essential for the dynamics reconstruction job particularly for time-delayed physical or biological  systems such as the gene regulation model above. Thus, the proposed time-delayed RC framework has a more flexible capability to deal with dynamics reconstruction jobs requiring tunable MC.

Finally, to further validate the efficacy of the time-delayed RC in reconstructing high-dimensional spatial-temporal system, we consider the ideal storage cellular automation model (ISCAM) simulating heterocatalytic reaction-diffusion processes at metal surfaces \cite{dress2010ideal,dress2011dynamics}. Considering the extremely high dimension (the $100 \times 100$ grids yields $10000$ input dimension), it is a challenging job to reconstruct the chaotic spatial-temporal patterns. As shown in Fig.~\ref{fig4}, with the same reservoir output dimension, the time-delayed RC has almost the same reconstruction ability as the non-delay one.

Our framework uses a few hyper-parameters, such as  $d$, the effective reservoir dimension, and $\tau$, the time delay, which definitely affect RC's efficacy in system reconstruction.  In fact, the existing literature included some criteria for selecting such parameters in system reconstruction using delayed embedding theory.   We thus implement these criteria, the dimension test and the delayed mutual information (DMI), to determine $d$ and $\tau$. From a perspective of embedding, $d$ is only required to be larger than $2\cdot\textrm{dim}(\mathcal{M})$ while practically the box-counting dimension of the manifold $\mathcal{M}$ is usually very small, i.e., $\textrm{dim}(\mathcal{M})$ is between $2$ to $3$ \cite{mcguinness1983fractal,VISWANATH2004115} for the chaotic Lorenz attractor.
 However, to design an effective RC, $d$
is required to be moderately large (see all the examples above). This is probably due that, although the generic property in the embedding theory means open and dense in a topological sense, there are still degenerated situations in practice, particularly for randomly-generated networks (see Fig.~S1 in~\cite{SM}).
Moreover, to reveal the mechanism from representation to computation, the recent efforts used the universal approximation theory \cite{NN2020} and the DMD \cite{chaos2021} framework which further  demonstrate the necessity of a large network size of RC in achieving good approximations.   Thus, the dimension tests are used to seek a suitable $d$ for each computation.  As for the delay $\tau$, either  too small or too large value renders computation problematic in system reconstruction, which naturally prompts us to introduce a modified DMI test taking into account the intrinsic time-scales of the neuronal dynamics in RC. Finally, it is noted that, for chaotic systems, the lagged observables earlier than the Lyapunov time have diminishing predictive power for the current time step, so we suggest the constraint $\tau\cdot\Delta t \cdot N_{\rm lag}<\Lambda_{\max}$ for the choice of $\tau$ and $N_{\rm lag}$ in practice where $\Delta t$ is the sampling stepsize. The details for the choice of these hyper-parameters are referred to \cite{SM}.

\begin{figure}
    \centering
    \includegraphics[width=0.5\textwidth]{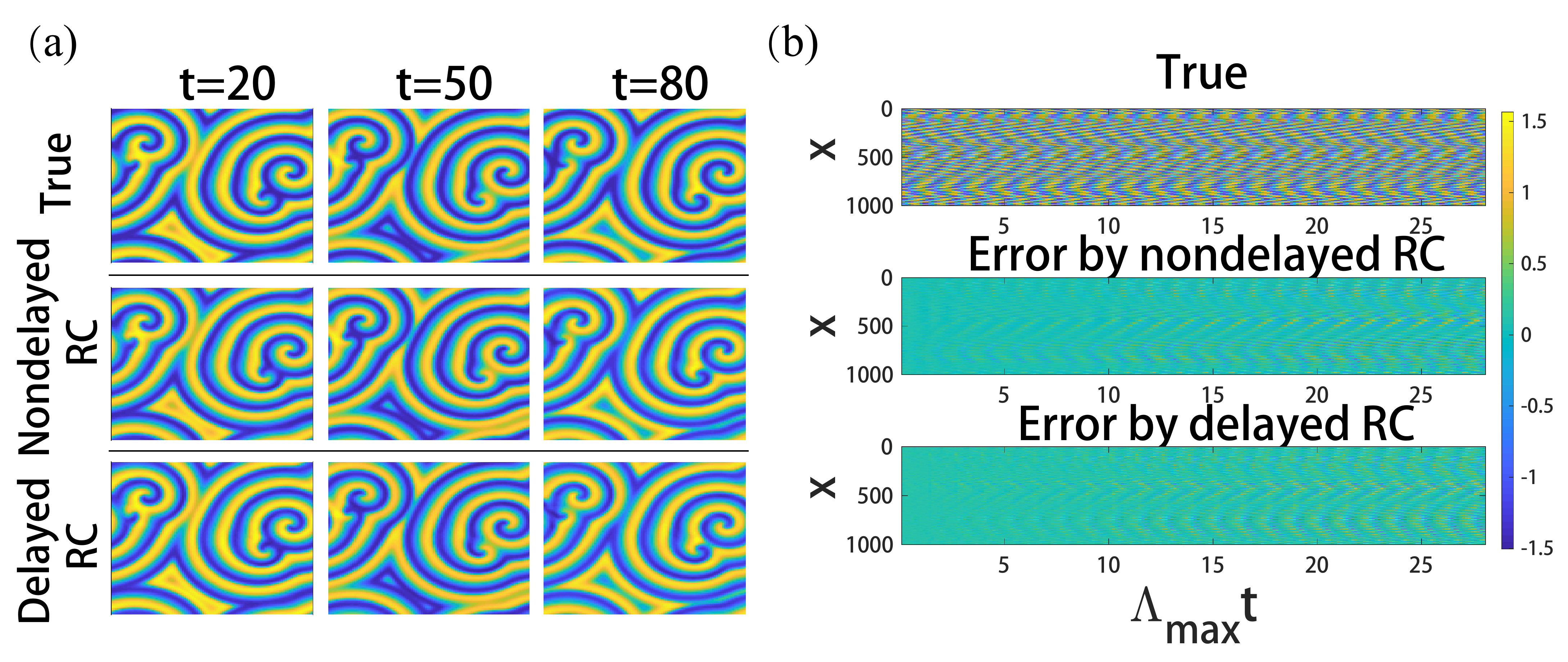}
    \caption{{ Reconstructed chaotic patterns for the ISCAM  model by a standard non-delayed RC including 5000 neurons and a time-delayed RC including 1000 neurons and 5 lags for each neuron. (a) Selected dynamical pattern using different evolution rules.  (b) Reconstruction errors deviating from true dynamics from different RC frameworks.  } }
    \label{fig4}
\end{figure}

{In conclusion, we have provided a deep and rigorous insight to the mechanism of RC from a viewpoint of embedding theory and nonlinear dynamical systems.  Based on our analytical findings, we have studied the role of time delay in the reservoir network and proposed a new framework of time-delayed RC.  This framework can significantly reduce the network size and promote the memory capacity, making its ability attain or even transcend the ability owned by the standard RC. Considering the computational costs which are crucially dependent on the network size in the dynamical evolution of RC and the hardware costs related to the circuit size in those overwhelmingly-developed physical RCs \cite{tanaka2019}, smaller-size reservoir is always expected to promote its real and extensive applications. 
Moreover, we notice a recently-published and independent work \cite{sakemi2020model}, where a method, different from the perspective of embedding theory and memory capacity presented here, was proposed to concatenating internal states through time in RC and realize model-size reduction.   Lastly, any contributions to designing RC frameworks of low-resource-consumption are believed to advance the direction of machine learning and thus be of broad applicability in solving data-driven science and engineering problems.}

This work is supported by the National Natural Science Foundation of China (Grant nos. 11925103 and 12171350), and by the STCSM (Grant nos. 18DZ1201000 and 2021SHZDZX0103).

\bibliography{ms}

\end{document}